\documentclass[conference, letterpaper]{IEEEtran}
\IEEEoverridecommandlockouts
\usepackage{cite}
\usepackage{amsmath,amssymb,amsfonts}
\usepackage{algorithmic}
\usepackage{graphicx}
\usepackage{textcomp}
\usepackage{xcolor}
\usepackage{geometry} % default margins
\usepackage{float}

\usepackage[caption=false,font=footnotesize]{subfig}

\def\BibTeX{{\rm B\kern-.05em{\sc i\kern-.025em b}\kern-.08em
    T\kern-.1667em\lower.7ex\hbox{E}\kern-.125emX}}

\begin{document}

\newgeometry{top=0.75in, bottom=0.77in, left=0.75in, right=0.75in}

\title{
\vspace{10pt}
Anisotropic Diffusion-Driven Ergodic Coverage in Multi-Robot Systems
}

\author{
\IEEEauthorblockN{Thales C. Silva}
\IEEEauthorblockA{\textit{Department of Computer Science} \\
\textit{Brown University}\\
Providence-RI, USA \\
thales\_silva@brown.edu}
\and
\IEEEauthorblockN{Anoop Kiran}
\IEEEauthorblockA{\textit{Department {of Computer Science}} \\
\textit{Brown University}\\
Providence-RI, USA \\
anoop\_kiran@brown.edu}
\and
\IEEEauthorblockN{Nora Ayanian}
\IEEEauthorblockA{\textit{Department of Computer Science} \\
\textit{Brown University}\\
Providence-RI, USA \\
nora\_ayanian@brown.edu}
\thanks{This work was supported by the
Office of Naval Research (ONR) Award No. N00014-24-1-2662 and the Brown University Seed Award from the Office of the Vice President for Research. Anoop Kiran is supported by the NSF Graduate Research Fellowship (Award No. 2040433).
Accepted for publication in the Proceedings of the IEEE International Symposium on Multi-Robot and Multi-Agent Systems (MRS), 2026. 
Final version available at https://doi.org/10.1109/MRS66243.2025.11357259}}

\maketitle

\begin{abstract}
We consider the problem of combining potential field and ergodic search on multi-robot systems. Traditional ergodic search algorithms use metrics for ergodicity that account for the desired distribution at different scales. Recently, a heat equation-driven ergodic approach was proposed, which adds flexibility to the smoothing of the ergodic metric. However, such an approach, as it is an isotropic diffusion, propagates the error uniformly in all directions, regardless of changes in the desired distribution. We introduce a general class of anisotropic diffusion formulation of the ergodicity problem, which generates a potential field for the ergodic search. We demonstrate that this approach generalizes previous results, which consider radial basis functions and the solution of the heat equation to represent the difference between the goal density distribution and the covered trajectories. In our solution, the agent movement is directed using the gradient of the solution of the Perona-Malik diffusion, and our formulation includes the heat equation as a special case. We demonstrate the methodology with a series of simulations in different scenarios.
%\textcolor{blue}{and also on an experiment using quadrotors.}
\end{abstract}


\vspace{-15pt}
\section{Introduction}
Multi-robot systems involve multiple robots interacting, coordinating, collaborating, or even competing to perform complex tasks. These systems have been used to accomplish a variety of tasks, from environmental monitoring \cite{Patel2021,Lynch2008,popa2005robotic,edwards2025distributed} to surveillance \cite{da2024communication,ji2022virtual}. In many of these tasks, the ability to efficiently search the environment is a fundamental part of the mission, be it for data-acquisition or search and rescue \cite{2022Kumar,baxter2007multi}. 
This paper considers the problem of coverage control in scenarios where a probability distribution of the target is known and proposes a solution that is based on diffusing the error between the target distribution and the distribution of the trajectories of the robots.

The coverage problem can be considered as an optimization problem with the task of finding an optimal distribution of robots over a desired target distribution. See \cite{thrun2002probabilistic} for foundations on the basic theory of optimal search. Within this view, there are two main types of coverage, namely static coverage, where the task is to find an optimal stationary configuration for the robots, and dynamic coverage, in which the robots move over the target distribution as their averaged trajectories approach the desired distribution.

A trajectory is ergodic with respect to a spatial distribution if the time spent in each region is proportional to that region's spatial density, ensuring thorough coverage without overlooking low-density areas.
Ergodic trajectories can be computed by various methods \cite{de2016ergodic,Mezic2017,ayvali2017ergodic,dressel2019tutorial}, with projection-based trajectory optimization being the most prevalent approach \cite{dressel2019tutorial,miller2013trajectory}. In such a method, a gradient descent is taken on an objective function that has ergodic cost and other costs of interest, such as energy or control effort.

Traditionally, potential field approaches are widely used for planning the trajectories of robots towards desired distributions because of their simplicity and elegance. Intuitively, it provides attractive forces that draw the robot toward goals, enabling smooth, feedback-driven trajectories. Early formulations include Khatib’s real-time obstacle avoidance using artificial potentials \cite{khatib1985} 
and navigation functions \cite{koditschek1990robot}. More recent work has addressed classical limitations: dynamic fractional-order repulsive fields incorporate velocity-dependent repulsion to improve responsiveness in moving environments 
\cite{ROBINET2024113}, while hybrid frameworks integrate global search methods like A* into the potential field to provide provable path optimality while retaining reactivity 
\cite{Chuanxiang2024}.
However, on its conventional formulation, potential fields do not account for regions where the desired density is relatively small and usually focus on exploitation, \textit{i.e.,} the robots are directed to the regions with the highest probability. Recently, such an approach was incorporated with ergodic search \cite{Mezic2017}, where the heat equation is used as a diffusing operator of the difference between the goal density distribution and the current coverage density.

An ergodic trajectory finds a balance between exploitation and exploration by design. It allocates more sampling time to regions that yield the most informative data while still maintaining broad spatial coverage, rather than concentrating solely on the highest-density points \cite{dressel2019tutorial,Miller2016}. By doing so, this approach enhances robustness to modeling errors through a more comprehensive and adaptable sensing strategy. There are numerous ways to generate an ergodic trajectory; for example, \cite{Sylvain2023,Mezic2017} formulate a heat equation that smooths the difference between the desired and actual coverage distributions in a multi-robot system, and then use the resulting potential field to steer each agent’s motion. In \cite{Mathew2011}, the robots perform the coverage of global features first and focus on small/local features later on. 

%\restoregeometry

To enhance the feature-preserving qualities of the ergodic search, we utilize the Perona-Malik diffusion \cite{PeronaMalik1990} to smooth the error of the potential field, allowing agents to be aware of structural edges in the distribution.
Structural edges in the search distribution mark sharp transitions between high-value and low-value regions, effectively outlining frontiers for targeted exploration. By preserving these boundaries, multi-robot teams can avoid redundant coverage, accurately map environmental features like soft obstacles or plumes, and more readily detect where sensor gradients are steepest. Moreover, as edges evolve--such as when a chemical plume drifts--robots can dynamically reassign themselves to follow shifting boundaries, ensuring that sensing remains thorough.
In short, Perona-Malik diffusion smoothly spreads coverage information across the spatial domain, but unlike classical smoothing methods, it adapts its diffusion strength based on local gradient magnitudes.

In this work, we generalize the results from Ivi\'c et al. \cite{Mezic2017} to consider a diffusion smoothing of the gradient of the coverage error, which allows for maintaining strong changes on features of the target distribution which could mean, for example, regions that interface distributions of interest and regions that the agents should avoid--but are not hard constraints such as obstacles. In particular, similar to that result, our methodology combines ideas from spectral multiscale coverage and potential field theory.
In a nutshell, the contributions of this work are twofold: first, we introduce a novel ergodic coverage strategy for multi-agent systems using the Perona-Malik anisotropic diffusion, which preserves spatial features and structural edges in the target distribution; and second, we demonstrate mathematically and numerically that our anisotropic formulation generalizes existing approaches.

The remainder of the paper is organized as follows: the robot team model and the problem formulation are described in Section \ref{sec:preliminaries}. Our methodology is presented in Section \ref{sec:methodology}, together with details on our implementation. Section \ref{sec:results} has experimental results and simulations. Finally, conclusions are in Section \ref{sec:conclusions}.

\vspace{-7pt}
\section{Preliminaries}
\vspace{-3pt}
\label{sec:preliminaries}
To simplify and focus on the problem of coverage, we assume that the robots move in a bounded domain $\Omega \subset \mathbb{R}^2$ and have integrator dynamics,
\vspace{-6pt}
\begin{equation}
\dot x_i = u_i,
\label{eq:dynamics}
\vspace{-4pt}
\end{equation}
for $i=1,...,N$, where $N$ is the number of robots, $x_i\in \mathbb{R}^2$ and $u_i\in\mathbb{R}^2$ are the robot's position and input, respectively. In general, such dynamics can be achieved through a low-level controller. Using more complex dynamics can be considered in our methodology, but its analysis is tangential to the contribution of this paper and it is left as future work.

Let $\mu:\Omega \rightarrow \mathbb{R}_{\geq 0}$ be a probability density, so that $\int_\Omega\mu(x)dx=1$, and is zero outside of $\Omega$. We say that a continuous trajectory $x_i(t)$ is ergodic with respect to $\mu$ if, for every measurable subset $S\subseteq \Omega$, the fraction of time that $x_i(t)$ spends inside $S$ converges to the mass of $S$ under $\mu$. In particular, define the empirical time-averaged measure of the trajectory on set $S$ up to time $T$ by
\vspace{-4pt}
\begin{equation}
\bar{d}_T(S) = \frac{1}{NT} \sum_{i=1}^N \int_0^T \boldsymbol{1}_{\{x_i(\tau)\in\mu(S)\}}d\tau,
\end{equation}
where $\boldsymbol{1}_{\{\cdot\}}$ is the indicator function.
Then, ergodicity means \cite{Petersen_1983},
\vspace{-6pt}
\begin{equation}
    \lim_{T\rightarrow \infty} \bar{d}_T(S) = \int_S \mu(y)dy,
\end{equation}
for every measurable $S\subseteq \Omega$. In particular, no region $S$ is over- or under-visited in the long run--each region is visited in proportion to its measure. Equivalently, for any continuous test function $\phi : \Omega \rightarrow \mathbb{R}$,
\vspace{-5pt}
\begin{equation}
    \lim_{T\rightarrow \infty} \frac{1}{NT}  \sum_{i=1}^N \int_0^T \phi(x_i(t))dt = \int_S \phi(y) \mu(y)dy.
\vspace{-3pt}
\end{equation}
In essence, because the trajectory visits each region in proportion to $\mu$, the time-average of any observable $\phi$ converges to its expected value under $\mu$. In this work, we propose a solution for the following problem.
\\
{\textbf{Problem 1:} 
Let $\Omega \subset \mathbb{R}^2$ be a bounded environment which has a spatial \textit{importance} field of interest $\phi(x)\geq 0$, \textit{e.g.}, chemical concentration. A team of $N$ mobile robots, each equipped with a local sensor, must collectively sample $\phi$ in $\Omega$, while spending more effort in regions where $\phi$ is large, without completely ignoring regions with low $\phi$.
}

We approach Problem 1 with a centralized approach and distributed execution. At the high level, a global coverage criterion derived from the Perona-Malik-smoothed error field dictates the collective sampling priorities for the robot team. This criterion establishes how much sampling effort should be allocated to different regions of the domain to achieve ergodic coverage. At the lower level, robots independently perform local sensing to update the global estimate of the importance field but follow trajectories computed from this centralized, globally informed criterion. Thus, trajectory decisions are centralized, while the robots individually provide continuous local measurements to update and refine the global coverage specification in real-time.

\section{Methodology}
\label{sec:methodology}
To solve Problem 1, we define the instantaneous coverage error as,
\vspace{-1pt}
\begin{equation}
    e(S,t) = \bar{d}_t(S) - \int_S\mu(y)dy,
\end{equation}
for $S\subseteq \Omega$.
The instantaneous coverage is a scalar field of the spatial error distribution. Inspired by \cite{Mathew2011,Mezic2017}, we define the following $L_2$-norm as metric for ergodicity,
\begin{equation}
    \label{eq:global_error}
    E(t) = \lVert e(\Omega,t) \lVert_2.
\end{equation}
This is the global error between the time-averaged density of the robots’ trajectories and the desired distribution, that is, it quantifies how far the fraction of the time spent by the robots in $\bar d_T(S)$ is from the measure of the density in $\mu(S)$. In short, it is an estimate of the overall coverage. Therefore, a solution of Problem 1 in terms of \eqref{eq:global_error} can be stated as
\begin{equation}
    \label{eq:error_zero}
    \lim_{t\rightarrow \infty} E(t) = 0.
\end{equation}

As pointed out in \cite{Mezic2017}, in general, \eqref{eq:error_zero} cannot be made monotonically decreasing.
To handle this issue, Ivi\'c et al. proposed a coverage method in which the agents follow the gradient of the stationary heat equation as a smoothing operator
for the spatial error field, which propagates information  to insufficiently covered areas.
In this work, we propose a generalization of that approach, where the Perona-Malik diffusion is considered as a smoothing operator for the gradient error. We show that the heat equation can be casted as a special case of our formulation.

\subsection{Perona-Malik Anisotropic Diffusion}
The Perona-Malik anisotropic diffusion \cite{PeronaMalik1990} is governed by
\begin{equation}
    \label{eq:perona-malik}
    \frac{\partial g(x,\tau)}{\partial \tau} = \nabla\cdot[D(\lVert\nabla g\lVert)\nabla g],
\end{equation}
with inital condition $g(x, 0) = e(x,t)$, where $g(x,\tau):\Omega \rightarrow \mathbb{R}$, and  $D(\cdot)$ is the spatially adaptive diffusivity. In this work, we define $D(\cdot)$ as
\begin{equation}
    \label{eq:pm_diffusivity}
    D(\lVert\nabla g(x,t)\lVert) = \frac{1}{1+(\lVert\nabla g(x,t)\lVert/K)^2},
\end{equation}
where $K\in\mathbb{R}$ is a user-defined constant parameter, which allows for the adjustment of the degree to which the gradient of the coverage error is considered. Such a structure allows diffusivity of the flow of information primarily along smooth areas in the environment (areas with small gradients), while restricting diffusion across sharp edges (areas with large gradients). Therefore, it allows for the preservation of structural boundaries--letting the agents focus on areas where the probability of interest changes smoothly.

The potential field $g(x,\tau)$ at time $\tau$, as a solution of \eqref{eq:perona-malik}, can be expressed using a nonlinear diffusion operator $\mathcal{D}_\tau$:
\begin{equation*}
    g(x,\tau) = \mathcal{D}_\tau(e(x,t)),
\end{equation*}
where the nonlinear operator $\mathcal{D}_\tau$ denotes applying the Perona-Malik diffusion for a duration of $\tau$. The parameter $\tau$ can be chosen to control the amount of diffusion (smoothing) at each time step.

Our proposed smoothing operator in \eqref{eq:perona-malik} has an elegant intuition. From the form in \eqref{eq:pm_diffusivity}, note that in small gradient regions, \textit{e.g.,} $\lVert \nabla g \lVert \approx 0$, the diffusivity $D$ tends to $1$, causing strong smoothing that evenly spreads error information, which facilitates exploration in those directions. On the other hand, in regions of high gradient, \textit{e.g.,} $\lVert \nabla g \lVert \gg K$, we have that the diffusivity $D$ tends to $0$, creating barriers to diffusion, thus preserving edges and sharps structures in the spatial distribution of the error defined in \eqref{eq:global_error}. 
In fact, {by choosing the diffusivity function $D(\cdot)$ as constant, the Perona-Malik diffusion in \eqref{eq:perona-malik} is then governed by
\begin{equation*}
    \frac{\partial g}{\partial \tau} = \beta \nabla^2 g,
\end{equation*} 
which is the linear diffusion, and in the steady state, it is exactly the stationary heat equation used in \cite{Mezic2017}}. In other words, with a constant diffusion coefficient, the anisotropic diffusion equations reduce to the heat equation, which is equivalent to Gaussian blurring \cite{guidotti2009some}.

\subsection{Agent Control Law from the Diffused Potential Field}
 In this work, we chose the following control law for the \textit{i}th robot,
 \begin{equation}
    \label{eq:control_law}
     u_i = v_m \frac{\nabla g(\boldsymbol{x}(t), t)}{\lVert \nabla g(\boldsymbol{x}(t),t)\lVert}
 \end{equation}
where $v_m$ denote the maximum speed of the agents, and $g(\cdot)$ is the solution of the Perona-Malik diffusion as defined in \eqref{eq:perona-malik}.
Due to the anisotropic diffusion preserving critical edges, the gradient fields are structurally informative, guiding agents to navigate narrow paths and complex spatial topologies inherent in the target distribution.

Although in this work we chose the same constant speed for every agent, their heading at any point in time is determined by their current position and the gradient of $g$, which drives down the overall coverage error.  

\vspace{-3pt}
\subsection{Numerical Semi-Implicit Implementation}
\label{sec:num_solution}
In this section, we discuss the implementation of our main result. For numerical efficiency, we perform spatial differentiation and divergence via Fast Fourier Transforms (FFT), and update it semi-implicitly.

\begin{figure}[htbp]
\centerline{\includegraphics[scale=0.85]{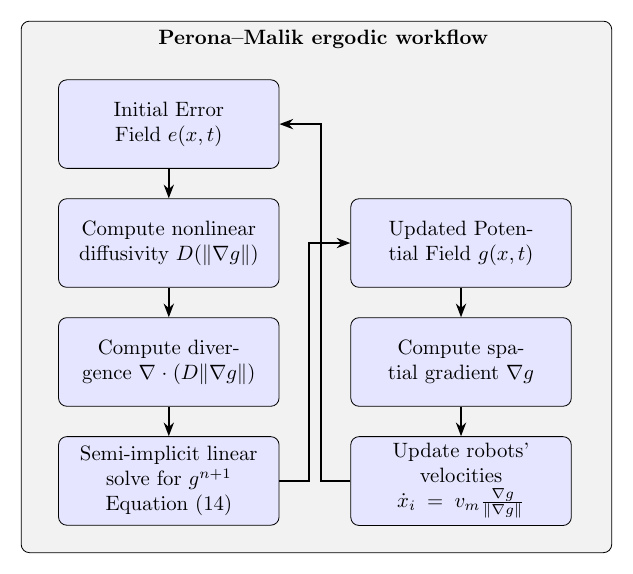}}
\caption{Schematic of the Perona-Malik anisotropic diffusion-based ergodic search.
Our method alternates between spatial and spectral domains. In the spatial domain, the nonlinear diffusivity is computed adaptively. The spectral domain efficiently computes spatial gradients, divergences, and performs a stable semi-implicit linear solve. Agents update positions using the gradient of the smoothed potential field.}
\label{fig:Schematic_figure1}
\end{figure}

Let $\hat g_k$ denote the Fourier coefficients of $g(x,t)$, then we compute the spatial gradients via FFT,
\begin{equation}
    \label{eq:fourier_spatial_grad}
    \nabla g(x,t) = \mathcal{F}^{-1}\{ik\hat g_k\},
\end{equation}
where $\mathcal{F}^{-1}\{\cdot\}$ stands for the inverse Fourier transform operator, $i$ is the imaginary number, and $k$ is the wave-number.
Next, we compute the nonlinear diffusivity $D(\lVert \nabla g \lVert)$ in the spatial domain and compute the divergence in the Fourier domain:
\begin{equation}
    \nabla \cdot (D(\lVert g \lVert )) = \mathcal{F}^{-1} \{ik \mathcal{F}\{D(\lVert \nabla g \lVert )\}\}.
\end{equation}
Finally, we update $g(x,\tau)$ using a semi-implicit Euler discretization. In particular, at each discrete time step,
\begin{equation}
    \label{eq:euler_semi_implicit}
    \frac{g^{n+1} - g^n}{\Delta t} = \nabla \cdot [D(\lVert \nabla g ^n\lVert)\nabla g^n] + \alpha \nabla^2 g^{n+1},
\end{equation}
where the additional linear diffusion term, $\alpha \nabla^2 g^{n+1}$, ensures numerical stability \cite{wickertRegularization}, where $\alpha \in \mathbb{R}$ is a constant. This yields a linear solution in the Fourier domain,
\begin{equation}
    \hat g_k^{n+1} = \frac{\hat g_k^n + \Delta t \hat{f}(g^n)_k}{1+\Delta t \alpha k^2},
\end{equation}
where $\hat{f}(g^n)_k$ is the Fourier transform of the nonlinear term at step $n$,
\begin{equation}
    f(g^n) = \nabla \cdot [D(\lVert \nabla g^n\lVert) \nabla g^n].
\end{equation}
This final step efficiently combines the spatially adaptive nonlinear smoothing with the spectral linear solve.
We compute adaptive nonlinear diffusion in physical space for accurate feature preservation, then immediately solve linear diffusion equations in the Fourier (spectral) domain for numerical stability. Our implementation is available in \cite{silva2025peronamalikcode}. A schematic of our approach is depicted in Figure \ref{fig:Schematic_figure1}.

\vspace{-15pt}

\begin{figure*}[ht]
  \centering
  \subfloat[Environment]{%
    \label{fig:sub0}%
    \includegraphics[width=0.23\linewidth]{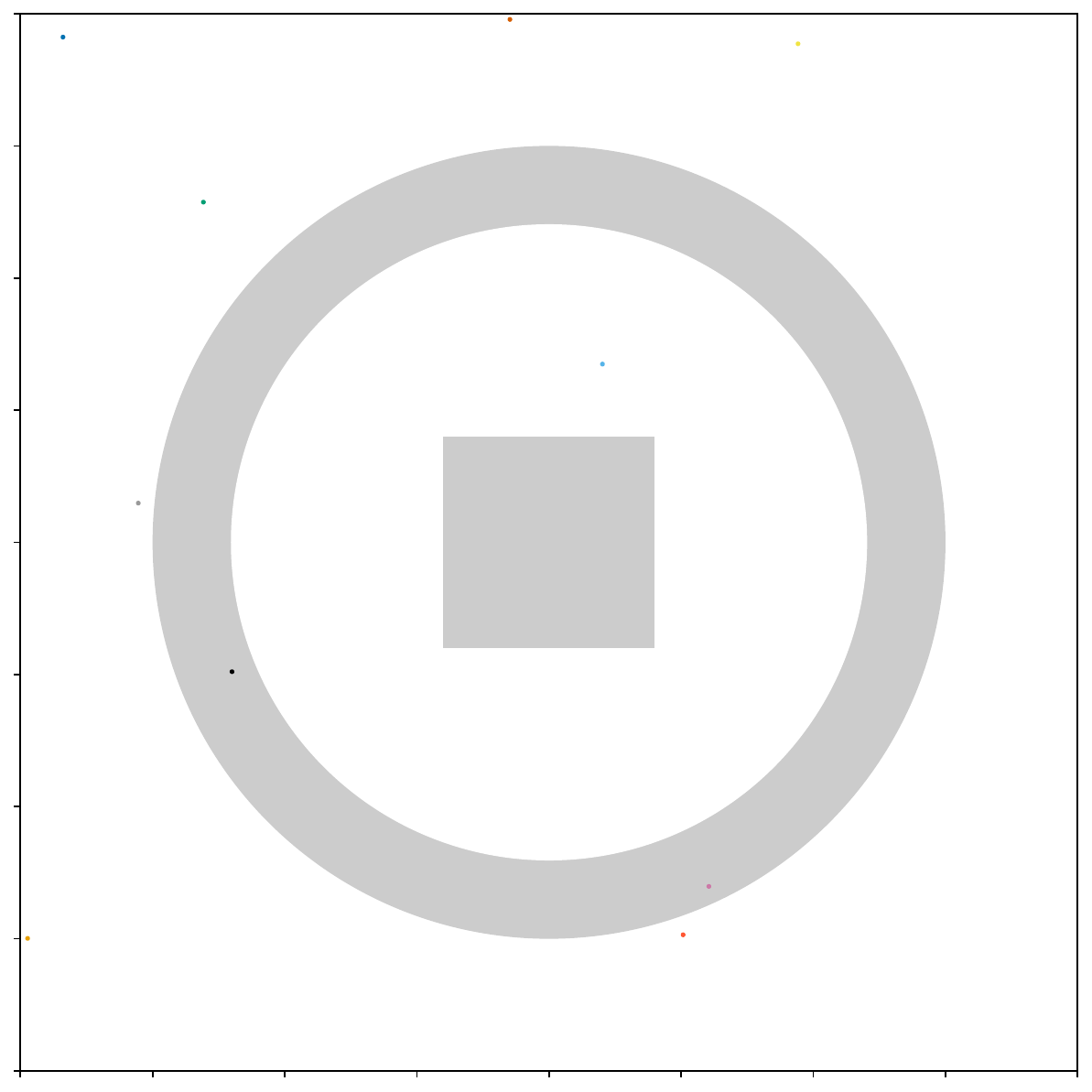}%
  }
  \subfloat[SMC]{%
    \label{fig:sub2}%
    \includegraphics[width=0.23\linewidth]{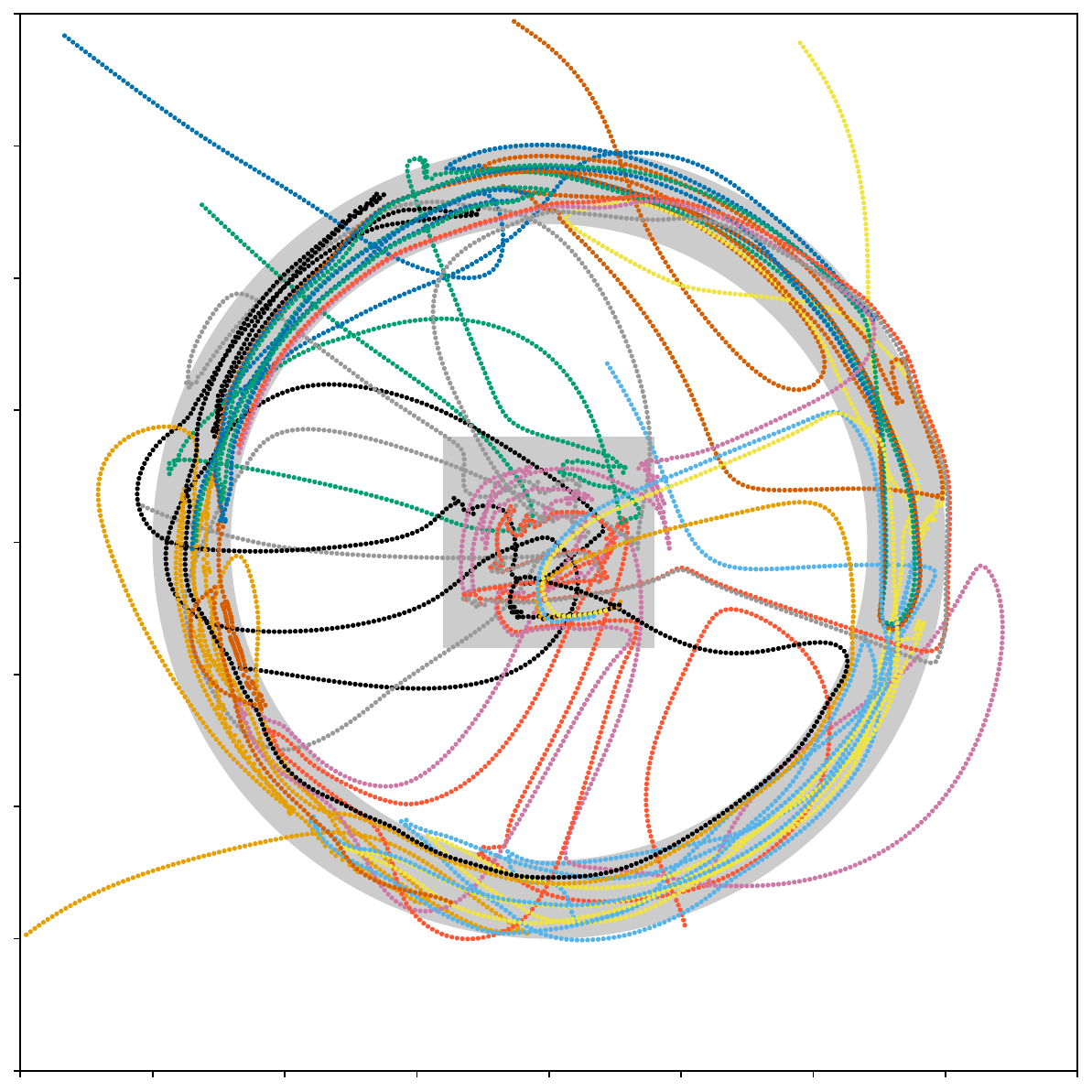}%
  }
  \subfloat[HEDAC]{%
    \label{fig:sub1}%
    \includegraphics[width=0.23\linewidth]{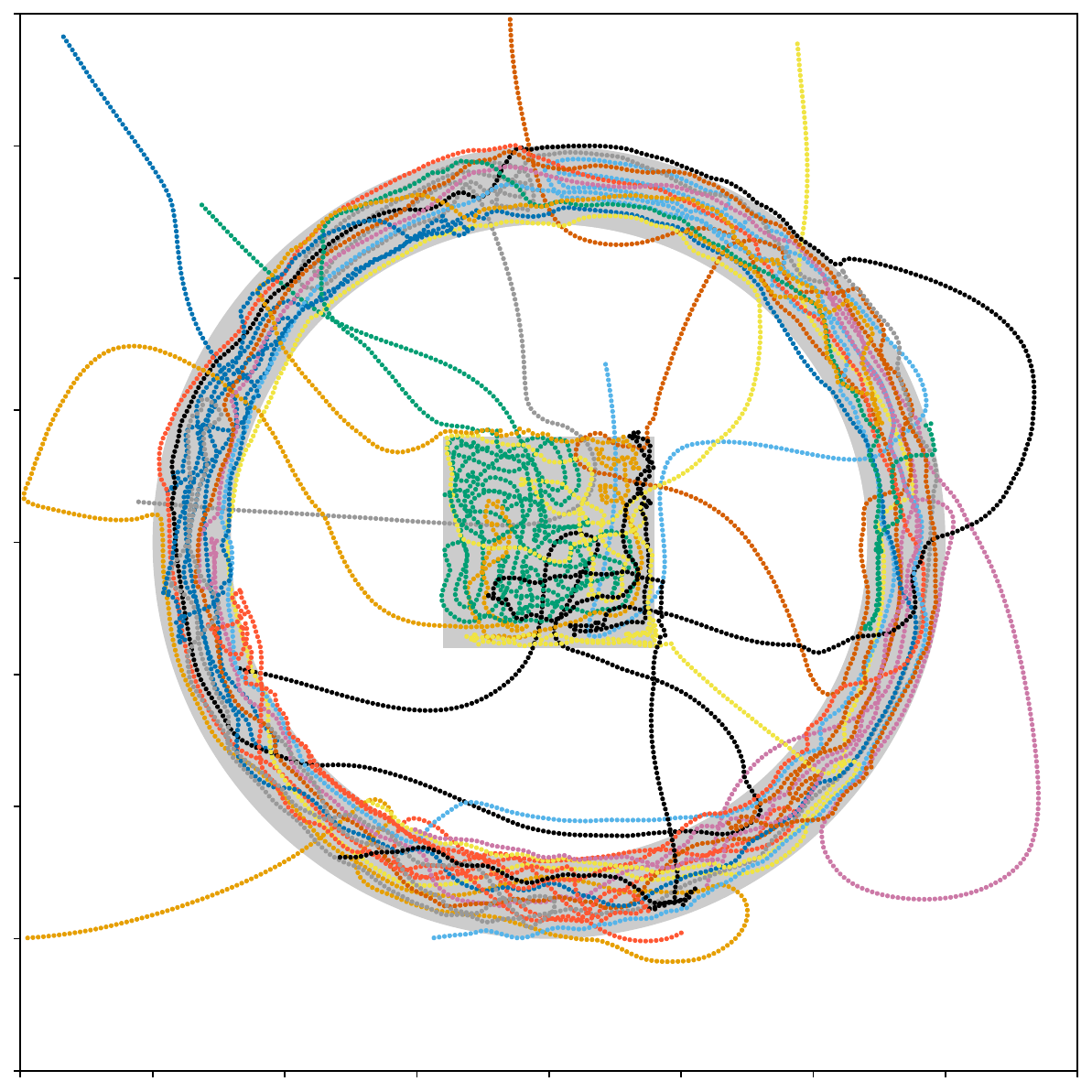}%
  }
  \subfloat[Perona-Malik (ours)]{%
    \label{fig:sub3}%
    \includegraphics[width=0.23\linewidth]{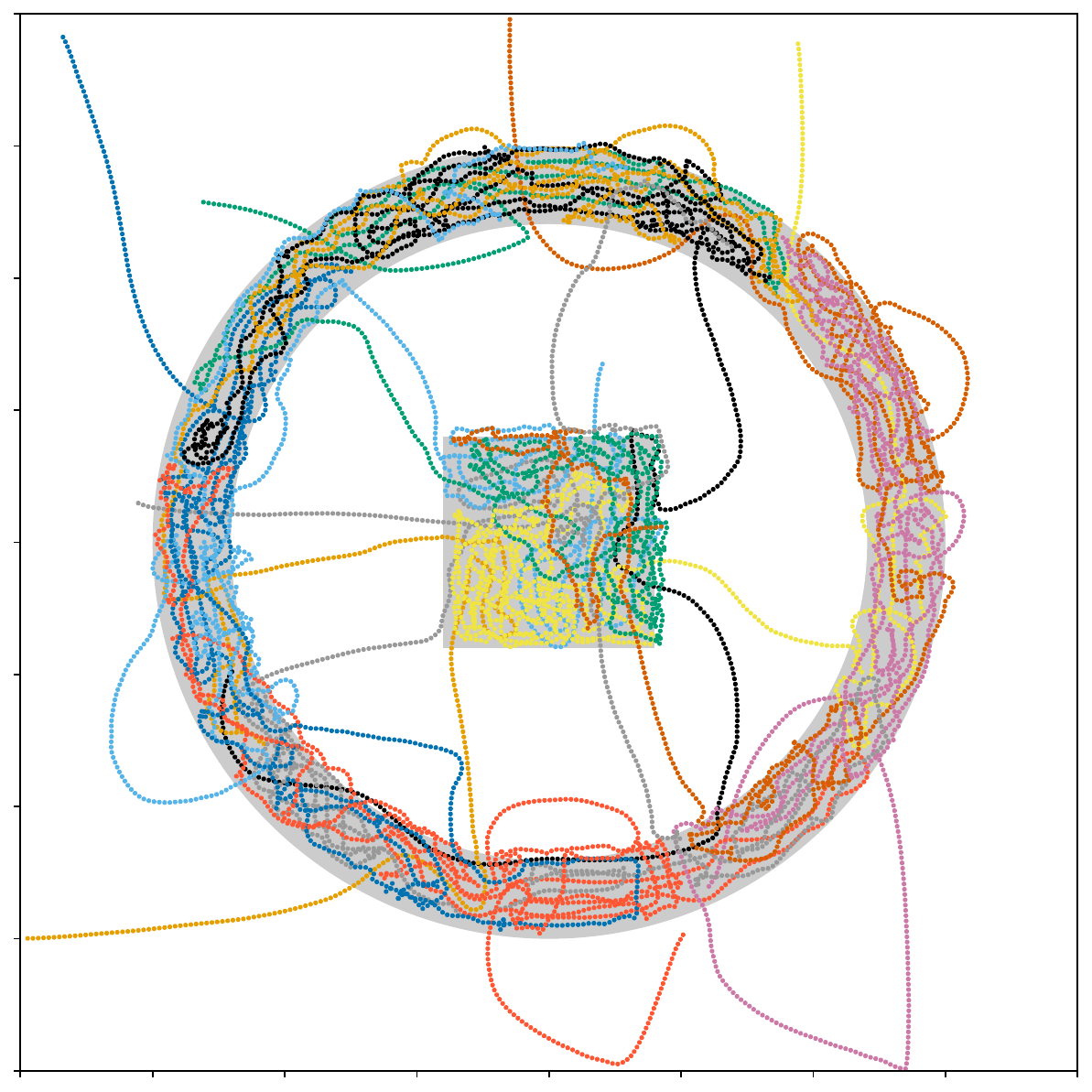}%
  }\hfil
  \\
  \subfloat[Environment]{%
    \label{fig:sub00}%
    \includegraphics[width=0.23\linewidth]{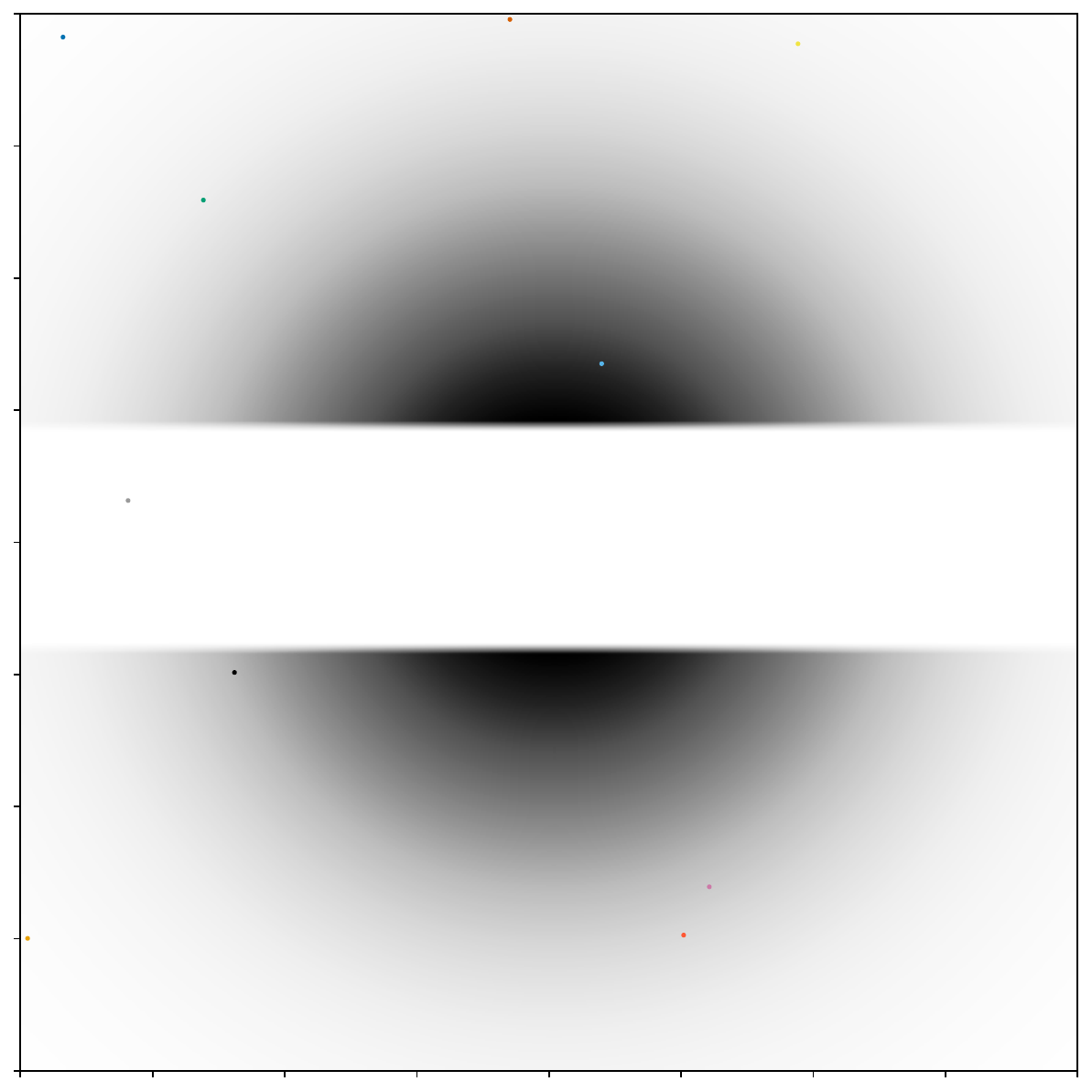}%
  }
  \subfloat[SMC]{%
    \label{fig:sub01}%
    \includegraphics[width=0.23\linewidth]{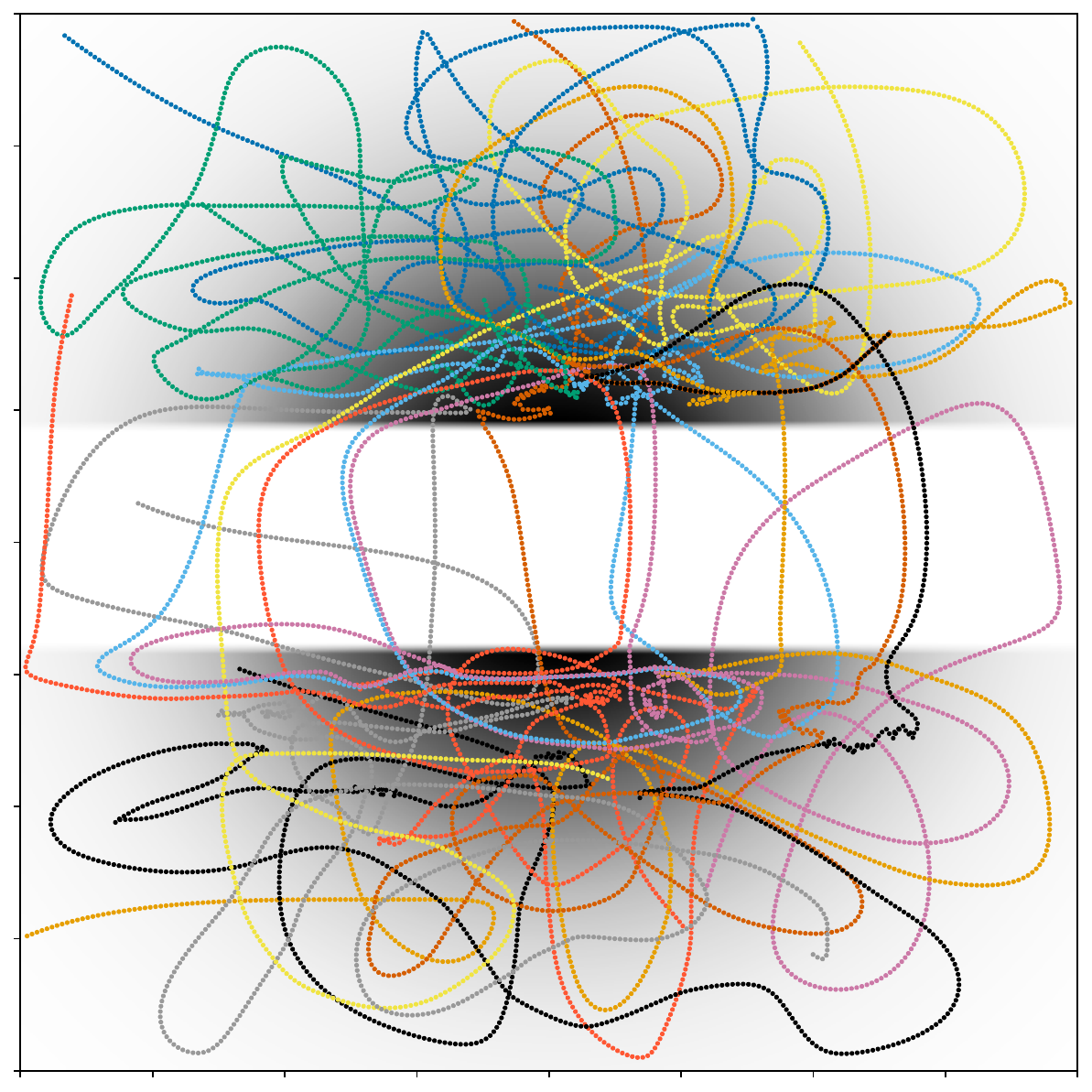}%
  }
  \subfloat[HEDAC]{%
    \label{fig:sub02}%
    \includegraphics[width=0.23\linewidth]{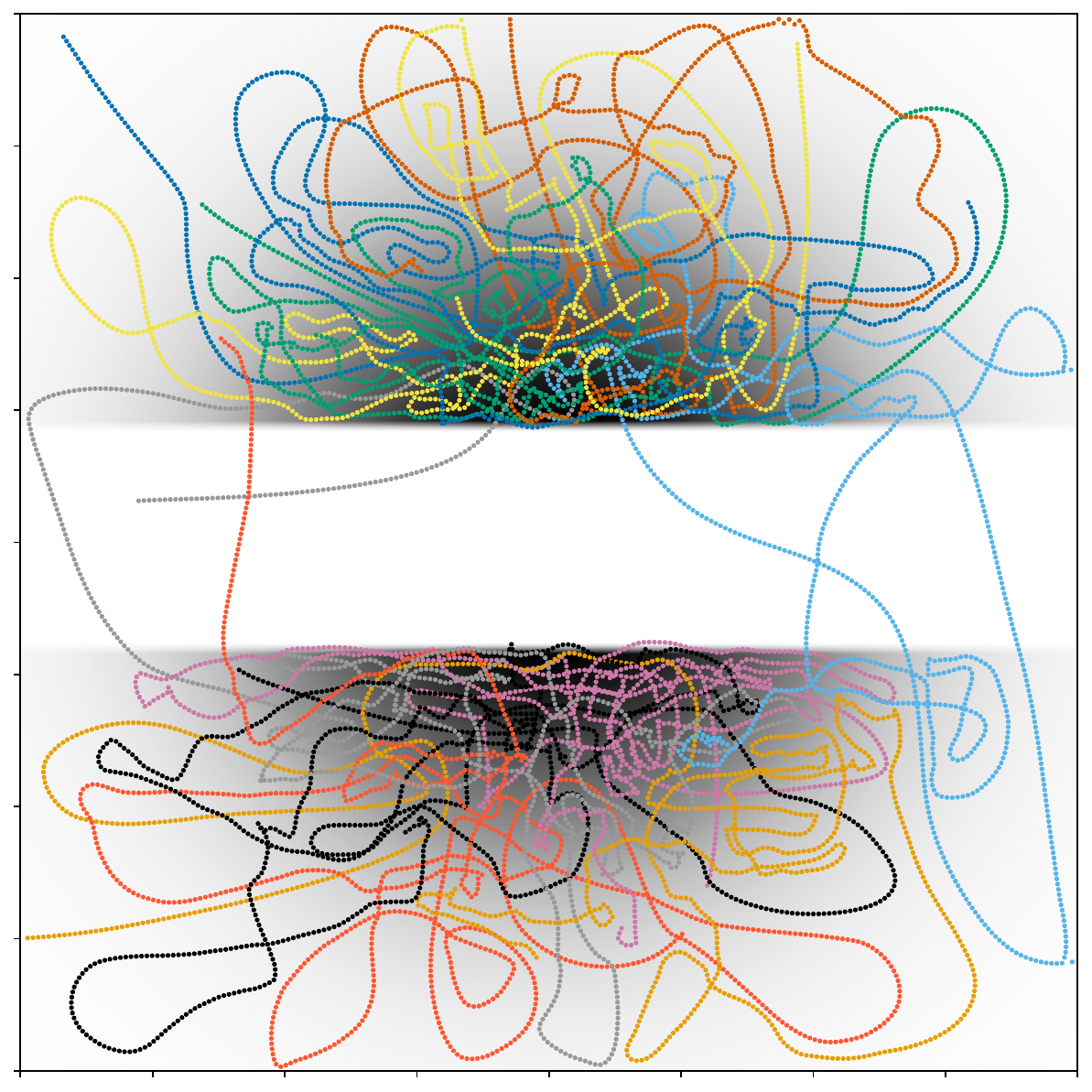}%
  }
  \subfloat[Perona-Malik (ours)]{%
    \label{fig:sub03}%
    \includegraphics[width=0.23\linewidth]{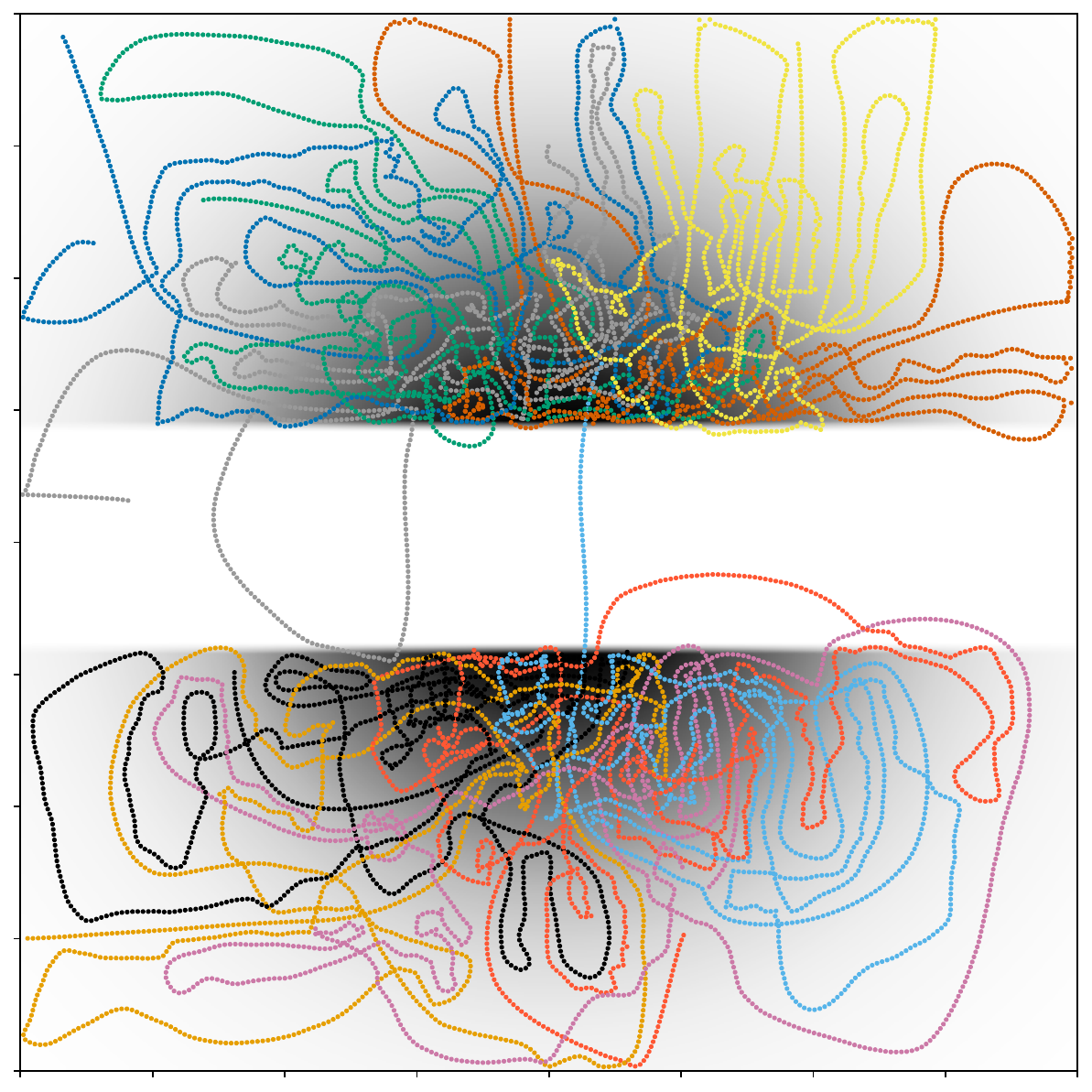}%
  }\hfil
  \\
  \subfloat[Environment]{%
    \label{fig:sub000}%
    \includegraphics[width=0.23\linewidth]{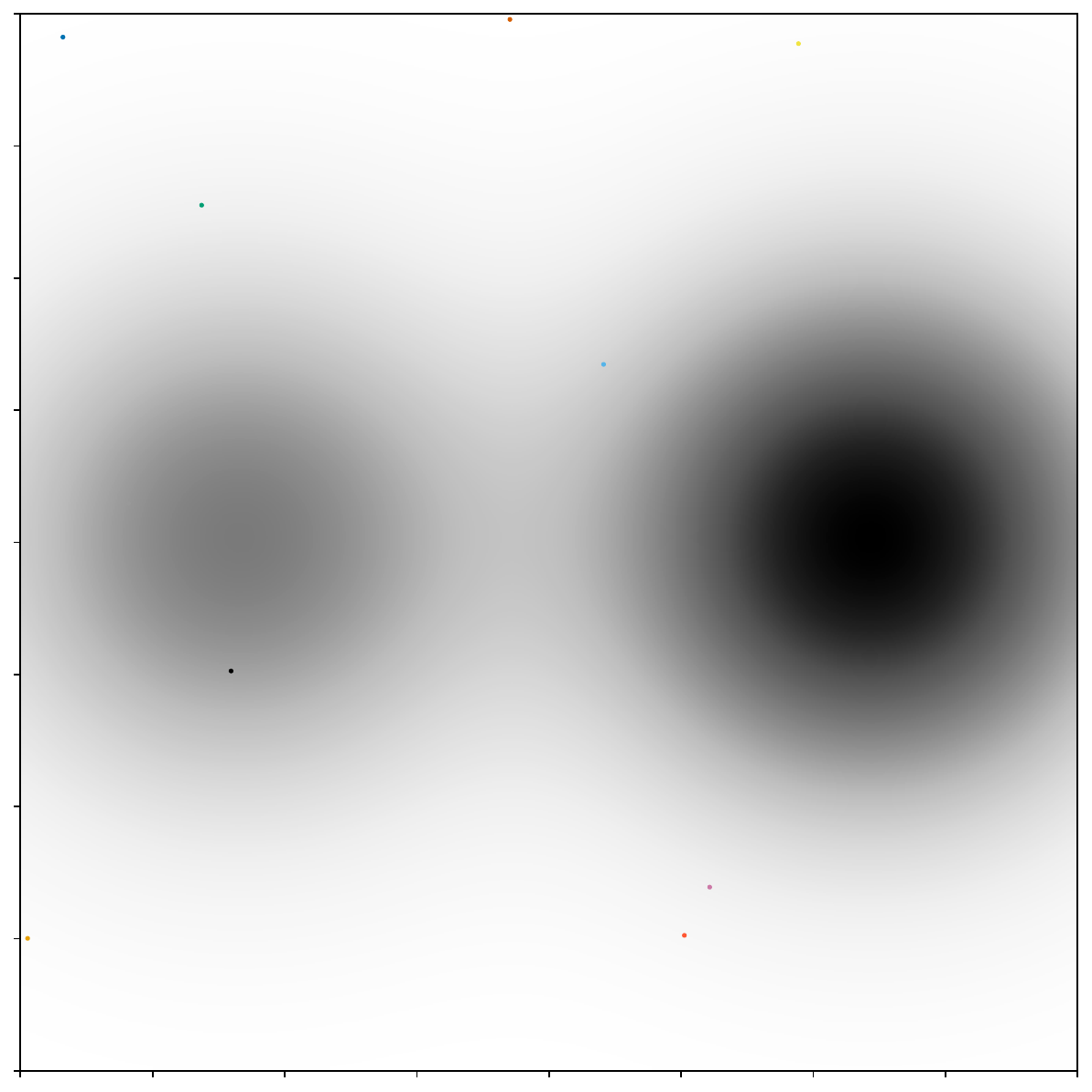}%
  }
  \subfloat[SMC]{%
    \label{fig:sub001}%
    \includegraphics[width=0.23\linewidth]{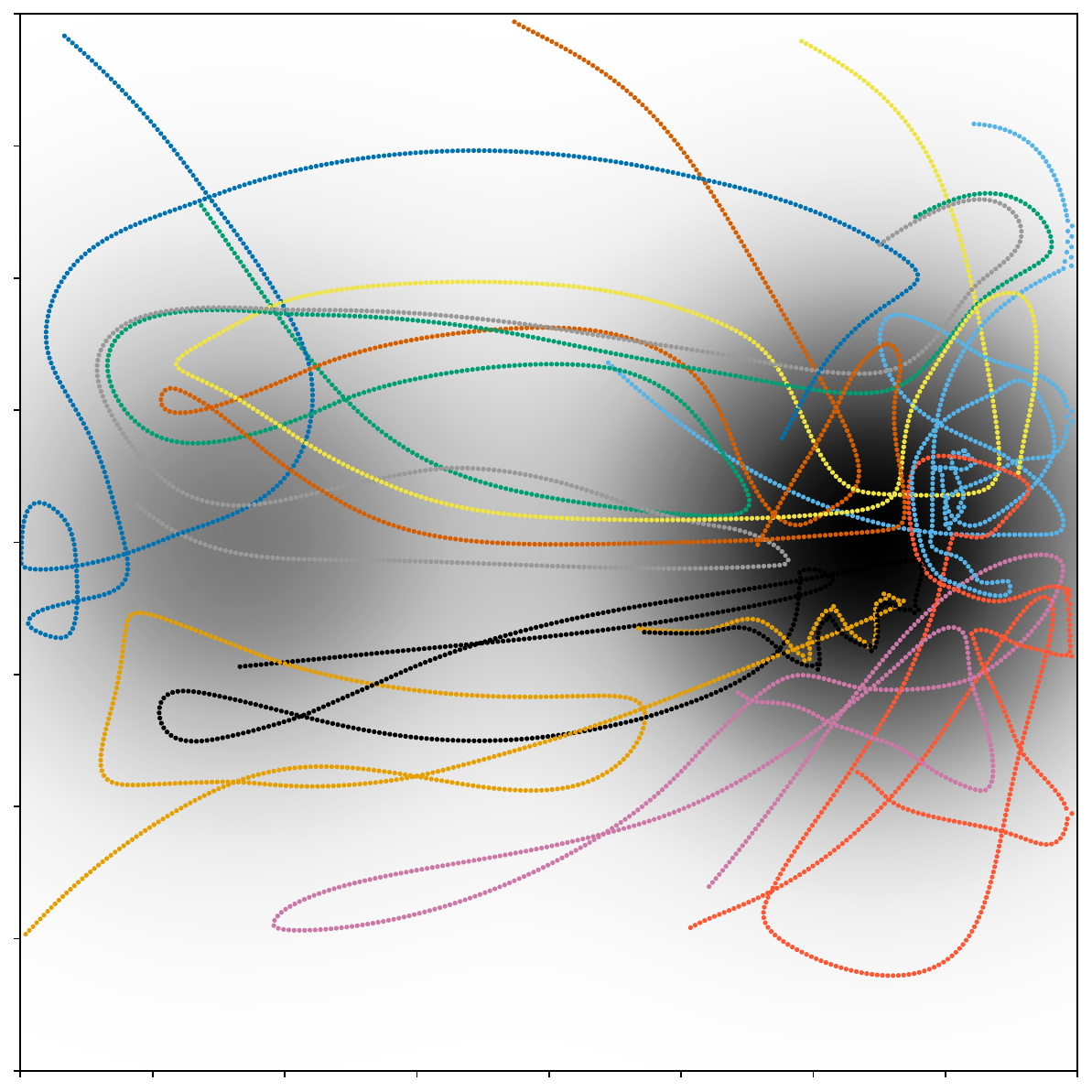}%
  }
  \subfloat[HEDAC]{%
    \label{fig:sub002}%
    \includegraphics[width=0.23\linewidth]{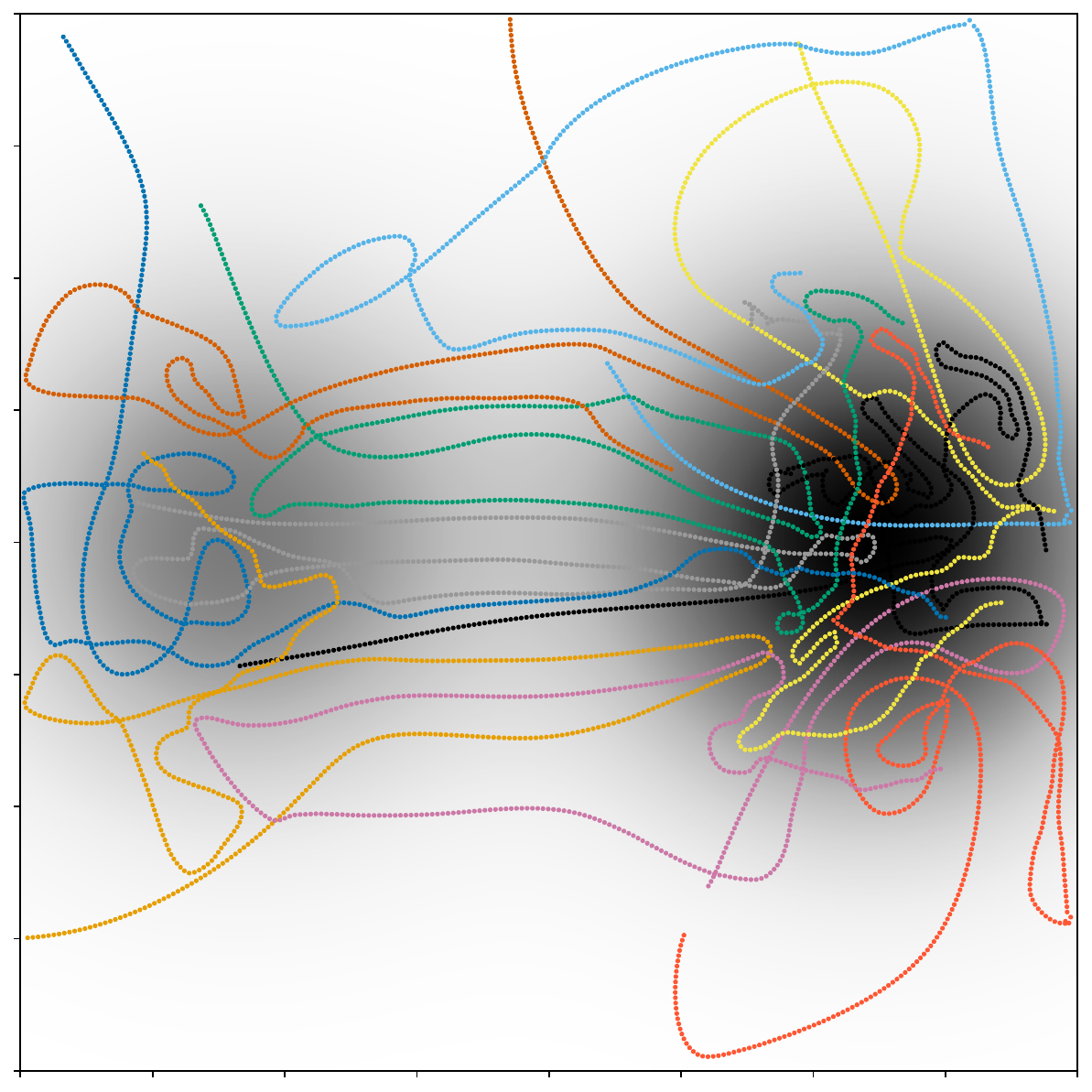}%
  }
  \subfloat[Perona-Malik (ours)]{%
    \label{fig:sub003}%
    \includegraphics[width=0.23\linewidth]{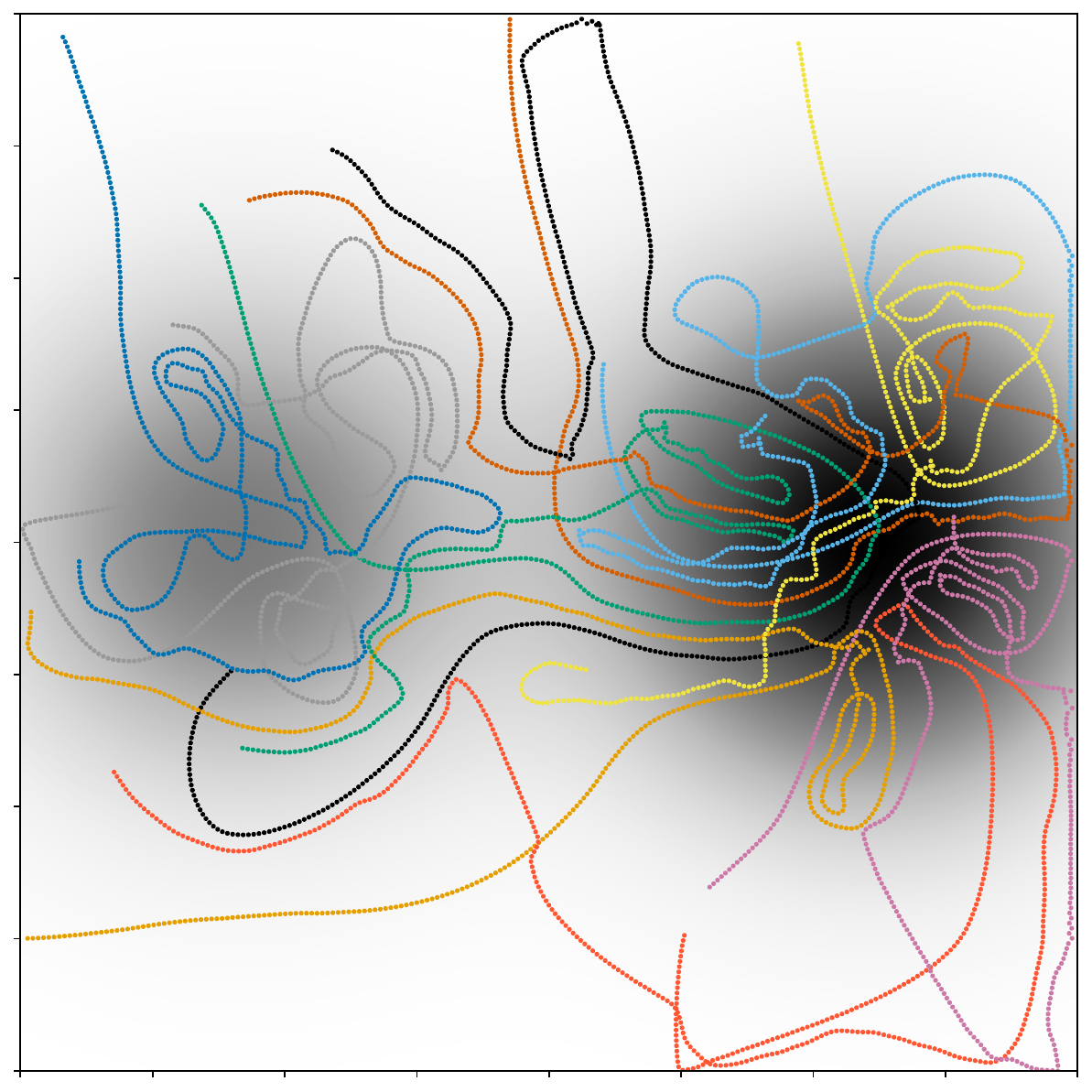}%
  }\hfil
  \caption{Figures (a)--(l) show one run with 10 robots and 1000 steps of each method--the HEDAC, SMC, and Perona-Malik (ours). The desired target distribution can be seen in Figures (a), (e), and (i) together with the robots' initial conditions. These figures illustrate that, with our method, the robots tend to cross areas with high changes in density fewer times than with the other approaches.}
  \label{fig:traj_all}
\vspace{-8pt}
\end{figure*}

\begin{figure*}[h]
  \centering
  \subfloat[Circle-Square]{%
    \label{fig:coverage_error_circle_square}%
    \includegraphics[width=0.30\linewidth]{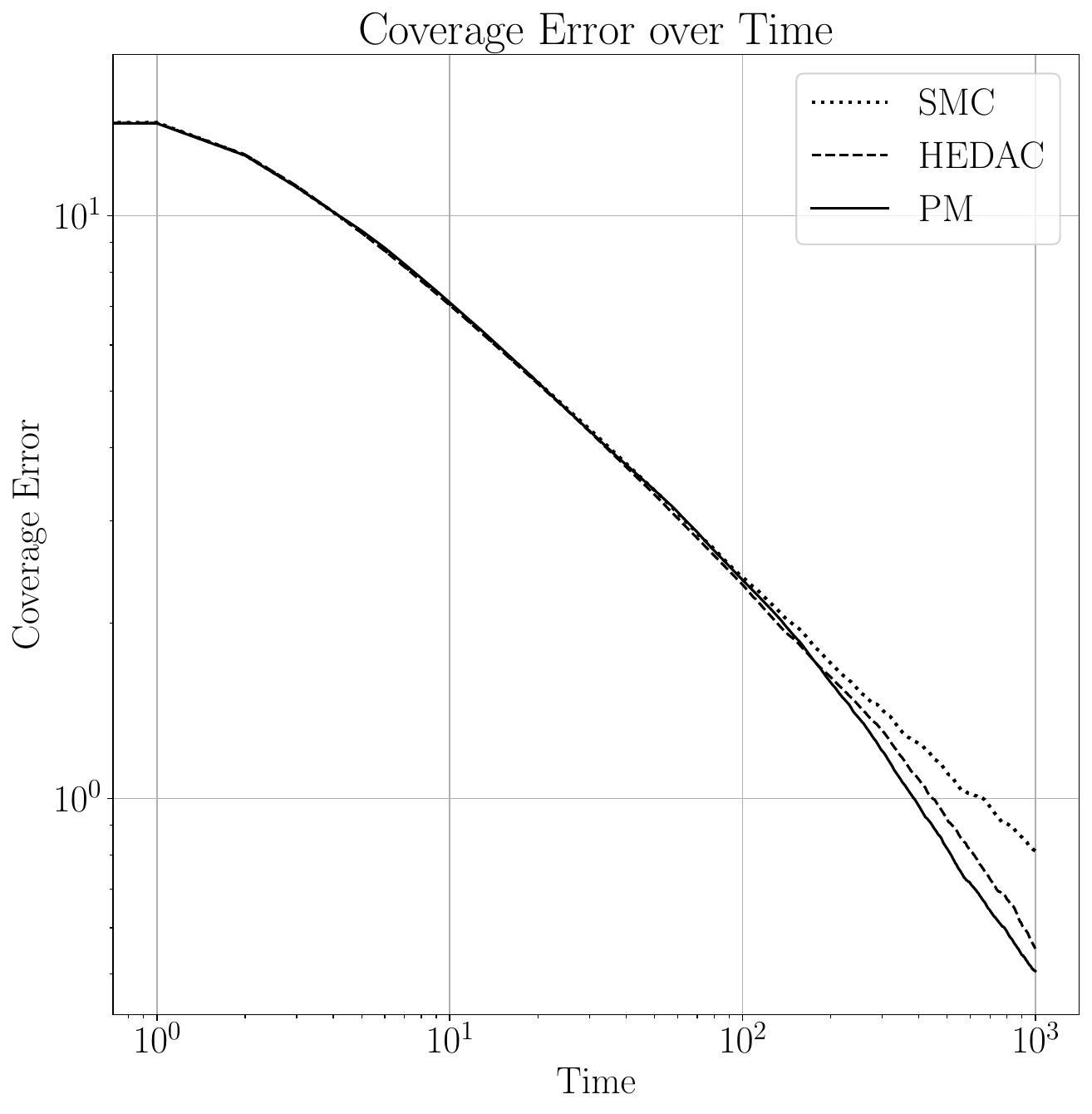}%
  }
  \subfloat[Gaussian with a stripe]{%
    \label{fig:coverage_error02}%
    \includegraphics[width=0.30\linewidth]{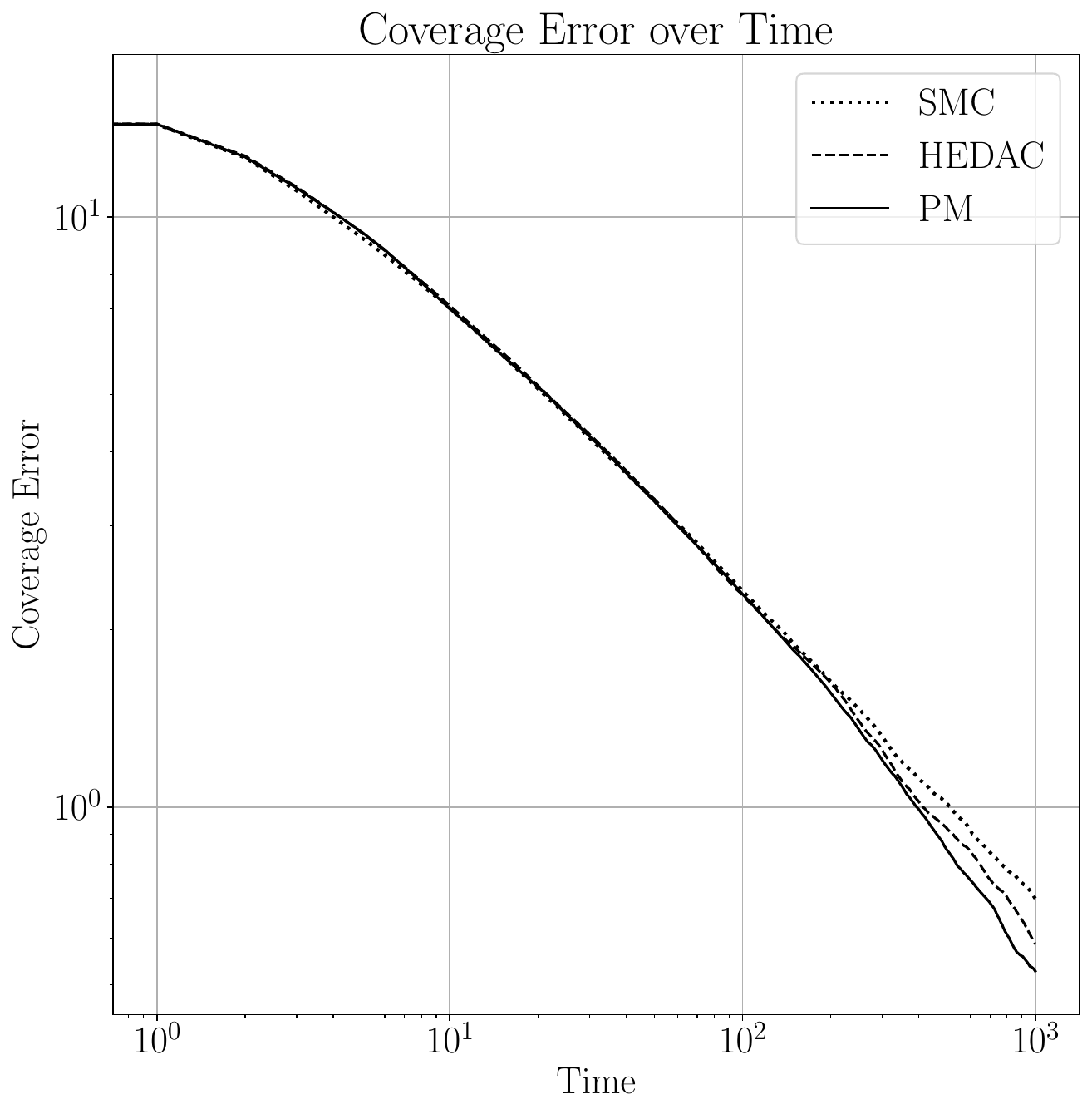}%
  }
  \subfloat[Bimodal Gaussian]{%
    \label{fig:coverage_error03}%
    \includegraphics[width=0.30\linewidth]{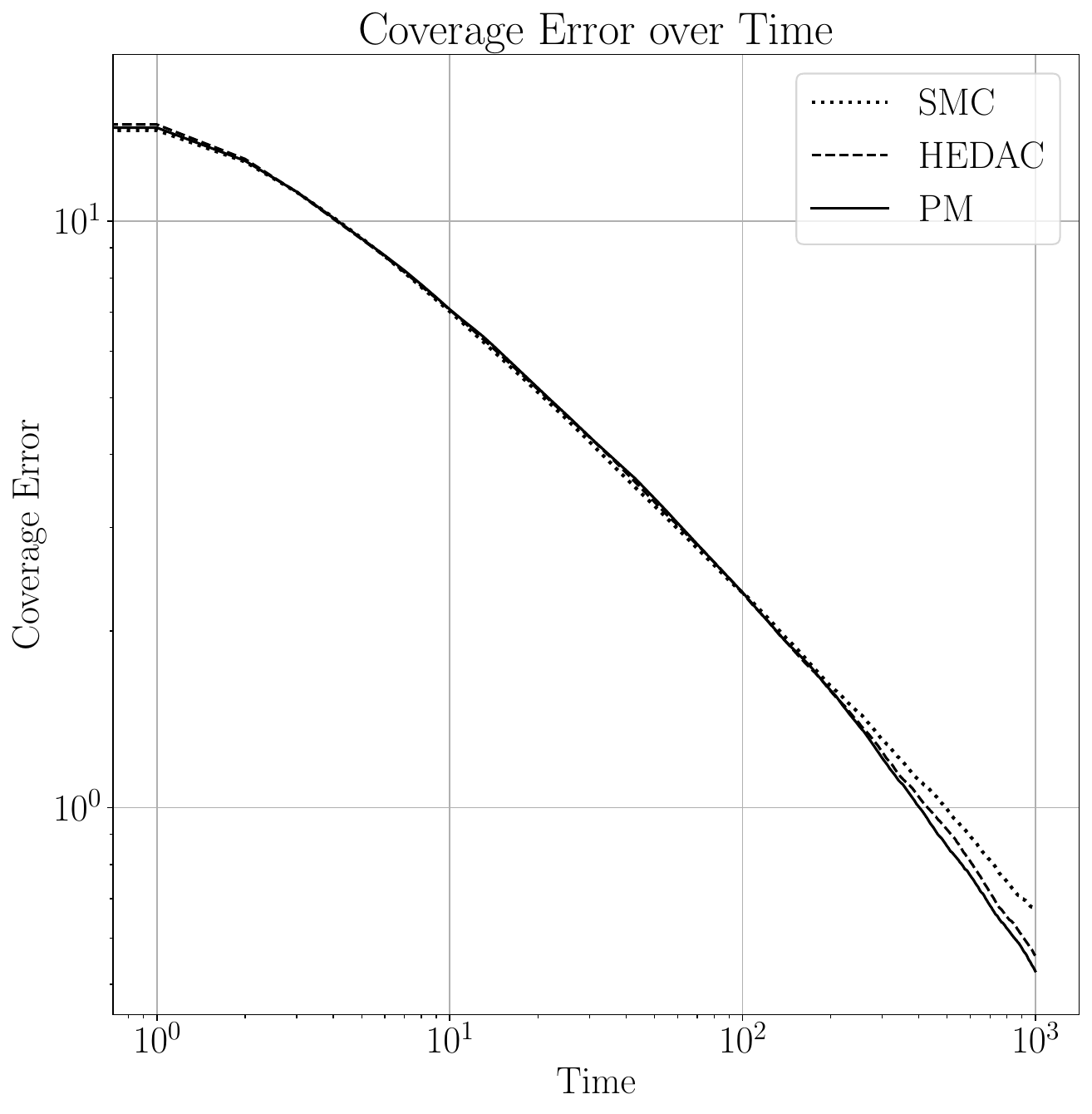}%
  }
\caption{Coverage errors of the environments in Figure \ref{fig:traj_all}, considering 50 runs with the same initial position between methods and runs. The coverage error is computed as equation \eqref{eq:global_error} for the whole environment. The result illustrates that the Perona-Malik approach for ergodic search converges faster than the SMC and the HEDAC. In addition, for environments where there are sharp changes in the target distribution, our approach has a better result than in scenarios where all the changes are smooth (\textit{e.g.}, the bimodal Gaussians.)}
\label{fig:coverage_error}
\vspace{-15pt}
\end{figure*}

\vspace{-0.35cm}
\section{Experiments and Results}
\label{sec:results}
In our simulation results, we aim to answer the following questions: (i) How effectively does the Perona-Malik diffusion preserve edge features in complex target distributions compared to traditional methods? and (ii) What performance improvements do the Perona-Malik diffusion provide in terms of quality and speed of total coverage?

To answer these questions, we conduct simulations comparing our proposed approach against traditional methods as baselines and analyze their performance across multiple scenarios.
As mentioned in the methodology section, our coverage algorithm is implemented over a 2-D domain with the Perona-Malik anisotropic diffusion model. Specifically, we discretize the Perona-Malik PDE in \eqref{eq:perona-malik}
on a uniform, orthogonal grid and employ the numerical method discussed in Section \ref{sec:num_solution}. Once the diffusion process yields the scalar field \(g\), we compute its spatial gradient in the spectral domain. 
In the numerical experiments that follow, agent trajectories are computed by integrating the motion equations with a forward Euler method.

We compare our method with the traditional spectral multiscale coverage (SMC) \cite{Mathew2011,mathew2010uniform,mavrommati2018} and with the heat equation driven coverage (HEDAC) \cite{Mezic2017}. To test our approach against those methods, we implemented the HEDAC, the SMC approaches, as well as ours, with the Dirac delta function to compute the robots' footprint coverage (please, note that using a radial basis function is also possible for all methods). We also show results with both, agents starting from the same random initial conditions, as well as robots starting from independent positions. For each test case, we show a total of 50 simulations and we show the coverage error, as defined in \eqref{eq:global_error}, in figures and tables.
For the SMC we used the traditional weights for the spectral frequencies as proposed in \cite{Mathew2011,mathew2010uniform}, while on the HEDAC we used normalized values for $\alpha$ and $\beta$, as suggested in \cite{Mezic2017}, and we set $\gamma=0$ since our method and the SMC do not have an intrinsic collision avoidance behavior. In our method, we used the following parameters during the simulations, $\Delta t = 0.05$, $K=0.1$, and $\alpha=0.5$; we also applied the diffusion for $\tau=0.9$ seconds. For the constant velocity, we set it to $v_m=1.0$. We notice that, while these parameters were used to maintain edges on the smoothed error between the robots' trajectories, averaged density, and the desired distribution, more aggressive values can be chosen to better preserve those structures. However, this can lead to very little diffusion in some cases, making the robots move erratically with constant velocity.

We test our algorithm in three different scenarios, one with mostly sharp transitions on the target density, one with sharp and smooth transitions, and another with only relatively smooth transitions. On the first test, with trajectories illustrated in Figures \ref{fig:sub0} to \ref{fig:sub3}, there are only sharp transitions between the square in the center and the circle around it. In all other regions, the target density is set to zero. In this case, we can see that our approach leads to fewer transitions between areas with zero density to areas with high density. The HEDAC algorithm has a similar behavior, albeit it has more jumps than ours, while SMC tends to smooth out the edges, and the robots tend to transition more often. This result is supported by the coverage error, illustrated in Figure \ref{fig:coverage_error_circle_square}, where our approach tends to converge faster than the other two. This is not surprising, given that our robots spend less time in areas of low interest.

Similar behavior can be noticed in the target density that is formed by a Gaussian with a zero density stripe in the middle, shown in Figures \ref{fig:sub00}, \ref{fig:sub03}, where our algorithm leads to fewer jumps in the region where there is a sharp transition. However, in this case, we also notice that our approach does not disregard all regions with low density--in this case, the regions where the transitions are smooth are well explored. Such a behavior is highlighted in the next environment, in Figures \ref{fig:sub000} to \ref{fig:sub003}. Here, as seen by the coverage error in Figure \ref{fig:coverage_error03}, the HEDAC and the Perona-Malik driven coverage have a closer performance. This can be explained by the fact that with regions with small gradients, the diffusivity term tends to be constant and causes even smoothing in all directions--facilitating the robots' exploration. 

%\subsection{Physical Experiments}
\vspace{-0.25cm}
\section{Conclusions}
\label{sec:conclusions}
In this paper, we have investigated the ergodic coverage problem for multi-robot systems in scenarios where a probability distribution of the target is known in advance. This problem is central to numerous practical applications, including search and rescue operations, environmental monitoring, and surveillance tasks. The ergodic search approach enables robots to explore environments efficiently by aiming to match the coverage distribution to the target distribution across multiple spatial scales. As a result, it naturally allocates more exploration effort to regions that contain valuable information, while maintaining sufficient coverage in lower-probability areas.

Specifically, our work extends prior literature by introducing anisotropic diffusion of the coverage error, allowing robots to perform searches that inherently respect structural edges and sharp transitions within the target distribution. Importantly, we have demonstrated mathematically that this anisotropic formulation generalizes both the heat equation-driven approach and radial basis function methods, encompassing them as special cases. Consequently, the anisotropic approach provides more efficient and targeted exploration by preserving essential spatial features of the target distribution. This improved efficiency arises primarily because agents actively avoid crossing high-gradient regions, which typically are boundaries between areas of contrasting importance. Through numerical examples presented in Section \ref{sec:results}, we have shown that this behavior can be effectively controlled and fine-tuned using parameters within our semi-implicit numerical implementation.

Several promising directions remain open for future work. We plan to formally analyze the convergence properties under time-varying distributions and identify explicit bounds on allowable parameter variation rates, extending our approach toward adaptive scenarios. Additionally, developing a completely distributed implementation of our ergodic coverage strategy represents another extension. Finally, a detailed assessment of the computational cost of our proposed method compared to existing techniques will provide insights and guidance for practical deployment.
\vspace{-5pt}
\bibliographystyle{IEEEtran}
\bibliography{library}

\end{document}